\documentclass{article}

\usepackage[final, nonatbib]{nips_2017}

\usepackage[utf8]{inputenc} 
\usepackage[T1]{fontenc}  
\usepackage{hyperref}     
\usepackage{url}            
\usepackage{booktabs}       
\usepackage{amsfonts}   
\usepackage{nicefrac}      
\usepackage{microtype}

\usepackage{amsmath}
\usepackage{bm}
\usepackage{caption}
\usepackage{subcaption}
\usepackage{enumerate}
\usepackage{mathtools}
\usepackage{url}
\usepackage[ruled]{algorithm2e}
\usepackage{hyperref}
\usepackage{color}
\usepackage{amssymb}
\usepackage{color}
\usepackage{comment}
\usepackage{enumitem}

\title{Multiple Adaptive Bayesian Linear Regression for\\ Scalable Bayesian Optimization with Warm Start}

\author{
  Valerio Perrone, Rodolphe Jenatton, Matthias Seeger, C\'{e}dric Archambeau \\
  Amazon\\
  Berlin, Germany\\
  \texttt{ \{vperrone, jenatton, matthis, cedrica\} @amazon.com} \\
}

\def\cb{{\mathbf c}}

\def\xb{{\mathbf x}}

\def\zb{{\mathbf z}}
\def\bb{{\mathbf b}}

\def\wb{{\mathbf w}}

\def\yb{{\mathbf y}}

\def\mub{{\boldsymbol\mu}}

\def\phib{{\boldsymbol\phi}}

\def\oneb{{\mathbf 1}}
\def\zerob{{\mathbf 0}}

\def\Xb{{\mathbf X}}
\def\Lb{{\mathbf L}}

\def\Ib{{\mathbf I}}

\def\Ub{{\mathbf U}}

\def\Kb{{\mathbf K}}
\def\Zb{{\mathbf Z}}

\def\Phib{{\boldsymbol\Phi}}

\def\Real{{\mathbb{R}}}

\def\Ocal{\mathcal{O}}

\def\Ncal{\mathcal{N}}

\def\Dcal{\mathcal{D}}

\def\Ucal{\mathcal{U}}

\begin{document}

\maketitle

\begin{abstract}
Bayesian optimization (BO) is a model-based approach for gradient-free black-box function optimization. Typically, BO is powered by a Gaussian process (GP), whose algorithmic complexity is cubic in the number of evaluations. Hence, GP-based BO cannot leverage large amounts of past or related function evaluations, for example, to warm start the BO procedure. We develop a multiple adaptive Bayesian linear regression model as a scalable alternative whose complexity is linear in the number of observations. The multiple Bayesian linear regression models are coupled through a shared feedforward neural network, which learns a joint representation and transfers knowledge across machine learning problems.
\end{abstract}

\section{Introduction}
Bayesian optimization (BO) is a well-established methodology to optimize expensive black-box functions~\cite{Shahriari2016}.
It relies on a probabilistic model of an unknown target function $f(\xb)$, which is repeatedly queried until one runs out of budget (e.g., time).  The queries consist in evaluations of $f$ at hyperparameter configurations $\xb^1,\ldots,\xb^n$ selected according to an explore-exploit trade-off criterion (e.g., expected improvement). The hyperparameter configuration corresponding to the best query is then returned. One popular approach is to impose a Gaussian process (GP) prior over $f$ and, in light of the observed queries $f(\xb^1),\ldots,f(\xb^n)$, to compute the posterior GP. The GP model maintains posterior mean and posterior variance functions as required by the explore-exploit criterion.

Despite their flexibility, GPs scale cubically with the number of observations~\cite{Rasmussen2006}. Hence, they cannot be applied in situations where $f$ has been or can be queried a very large number of times. In this work, we are interested in such a setting as we would like to warm start BO by, e.g., transferring information obtained from previous runs of the BO routine, or learn across similar problems (e.g.,
a given classifier applied across different datasets~\cite{Bardenet2013,Yogatama2014,Feurer2015,Fusi2017,Golovin2017}),
which we will call \emph{tasks}.
To tackle the scalability limitation of GPs and ease transfer learning in BO,
we propose to fall back to adaptive Bayesian linear regression (BLR)~\cite{Bishop2006}, ABLR for short, which scales linearly with the number of observations and cubically in the dimension of the basis function expansion. 
Sparse GPs~\cite{McIntire2016} or multi-task GPs~\cite{Swersky2013} have been respectively developed to scale up GPs and make them suitable for multi-task learning.
ABLR offers a simple alternative combining the strengths of these two approaches. 

Our main contribution is to learn a suitable representation of a variety of tasks with a feedforward neural network (NN), provided it is fed with enough data. We consider conditionally independent task-specific BLR models, which share a NN that learns the basis expansion. We compare to random Fourier basis expansions~\cite{Rahimi2007} as they have already been successfully applied to BO~\cite{Hernandez-Lobato2017,Jenatton2017b}. While more scalable, they are less flexible in learning a useful representation.

Closest to our work is \cite{Snoek2015}, where BO is scaled up by replacing the GP with an ABLR model. The authors consider a single task setting, with a two-step inference procedure. First, they train the NN with a squared loss at the output layer to learn a maximum a posteriori estimate of the NN parameters. This requires evaluating a number of candidate queries to feed the NN training algorithm. They then fix the network architecture and replace the output layer by a BLR layer to run the BO routine. Instead, we \emph{jointly} learn the basis expansion, that is, the NN, and the task-specific BLR models in one go. Our objective function corresponds to a sum of log-marginal likelihood terms, each term corresponding to one of the underlying tasks.
As a result, in contrast with \cite{Snoek2015} who use the squared loss, we can handle heterogeneous signals, each having its own marginal likelihood. In this sense, we borrow the strength of the likelihood of multi-output GPs while maintaining the scalability of~\cite{Snoek2015}.

Another related model is presented in~\cite{Springenberg2016}. The authors propose Bayesian NNs to sample from the posterior over $f$, and add task-specific embeddings to the NN inputs to handle multiple tasks. While allowing for a principled treatment of uncertainties, fully Bayesian NNs are computationally more expensive and their training can be sensitive to the stochastic gradient MCMC hyperparameters. Our model allows for simpler inference and is more suitable for large scale deployment.

\section{Multiple Adaptive Bayesian Linear Regression Model}
\label{sec:model}
Consider $T$ tasks defined by a set of black-box target functions $\{f_t(\cdot)\}_{t=1}^T$ we would like to optimize. Let $\Dcal_t = \{(\xb^n_t, y^n_t) \}_{n=1}^{N_t}$ be the set of $N_t$ pairs of inputs and responses associated to task $t$. We further denote the stacked response vector associated to task $t$ by $\yb_t \in \Real^{N_t}$ and the corresponding stacked matrix of inputs by $\Xb_t\in\mathbb{R}^{N_t \times P}$.
We assume the task responses $\{\yb_t\}_{t=1}^{T}$ are drawn from independent BLR models conditioned on the shared feature map $\phib_\zb(\xb):\mathbb{R}^{P}\mapsto\mathbb{R}^{D}$, which is parametrized by $\zb$, and the residual noise parameters $\{\alpha_t\}_{t=1}^T$:
$$
\yb_t \mid  \Xb_t, \wb_t, \alpha_t, \zb \sim \Ncal (\Phib_\zb(\Xb_t) \wb_t, \alpha_t^{-1} \Ib_{n_t}) ,
$$
where $\Phib_\zb(\Xb_t) = [\phib_\zb(\xb^n_t)]_n  \in\mathbb{R}^{N_t\times D}$ is the feature matrix, $\wb_t \in \mathbb{R}^D$ a weight vector, and $\alpha_t\in\mathbb{R}^+$ a precision (i.e., inverse variance). To complete the model, we impose a zero-mean isotropic Gaussian prior on $\wb_t$ and denote its precision by $\beta_t\in\mathbb{R}^+$. 
In the remainder, we will use $\Phib_t$ for $\Phib_\zb(\Xb_t)$. 

\subsection{Posterior inference}
The posterior distribution over the weight parameters is analytically tractable in this model, as well as the predictive distribution, both of which are multivariate Gaussian distributions~\cite{Bishop2006}. The predictive mean and the predictive variance at a new input $\xb_t^*$ are respectively given by 
\begin{align}
\mub_t(\xb_t^*; \Dcal_t, \alpha_t, \beta_t, \zb) &= \frac{\alpha_t}{\beta_t}\phib_{\zb}(\xb_t^*)^\top \Kb_t^{-1} \Phib_t^\top \yb_t = \frac{\alpha_t}{\beta_t}   \cb_t^\top \Lb_t^{-1}  \phib(\xb_t^*) ,\label{eq: predictive mean}\\
\sigma_t^2 (\xb_t^*; \Dcal_t, \alpha_t, \beta_t, \zb) &= \frac{1}{\beta_t} \phib_{\zb}(\xb_t^*)^\top \Kb_t^{-1} \phib_{\zb}(\xb_t^*) + \frac{1}{\alpha_t} = \frac{1}{\beta_t} ||\Lb_t^{-1} \phib(\xb_t^*) ||^2 + \frac{1}{\alpha_t}, \label{eq: predictive var}
\end{align}
where $\Kb_t = \frac{\alpha_t}{\beta_t} \Phib_t^\top \Phib_t + \Ib_D$. The right hand side reformulations (\ref{eq: predictive mean}) and (\ref{eq: predictive var}) ensure numerical stability. They are obtained by decomposing $\Kb_t$ in terms of its Cholesky factor $\Kb_t= \Lb_t \Lb_t^\top$, so that $\phib(\xb_t^*)^\top \Kb_t^{-1} \phib(\xb_t^*) = ||\Lb_t^{-1} \phib(\xb_t^*) ||^2$ and $ \Kb_t^{-1} \Phib_t^\top \yb_t = \Lb_t^{-\top} \cb_t$ with $\cb_t= \Lb_t^{-1} \Phib_t^\top \yb_t$.

Each task-specific BLR depends on the hyperparameters $\alpha_t$ and $\beta_t$, as well as the set of hyperparameters $\zb$ defining the feature map. In particular, $\zb$ will represent the weights of a NN (see Section~\ref{sec:nn features}). We adopt an empirical Bayes approach and jointly learn all these hyperparameters by optimizing the marginal likelihood of the data~\cite{MacKay2003}. More specifically, we integrate out the model parameters $\{\wb_t\}_{t=1}^T$ and minimize the sum of the negative log-marginal likelihoods of each task:
\begin{equation}\label{eq:marginal_likelihood}
\rho\left(\zb, \{\alpha_t, \beta_t \}_{t=1}^T\right) = - \sum_{t=1}^T \left[
 \frac{N_t}{2} \log \alpha_t - \frac{\alpha_t}{2} \left( ||\yb_t||^2 - \frac{\alpha_t}{\beta_t}  ||\cb_t||^2 \right) - \sum_{i=1}^D \log (  [\Lb_t]_{ii}) \right].
\end{equation}

\subsection{Learning a joint representation with feedforward neural networks}
\label{sec:nn features}
We learn the nonlinear map $\phib_{\zb}(\xb)$ with a feedforward NN. For some input vector $\xb$, we consider the following $L$-layer feedforward transformation parametrized by the weight matrices $\{\Zb_l\}_{l=1}^L$:
$$
\phib_{\zb}(\xb) = a_L\left(  \Zb_L a_{L-1} \left(  \dots \Zb_2 a_1\left(  \Zb_1 \xb  \right) \dots   \right)  \right) .
$$
The parameter vector $\zb$ is a flattened version of the stacked weight matrices. In practice, $a_l$ are set as \texttt{tanh} functions and $L=3$ (as~\cite{Snoek2015}), but any more complex NN architecture can be used.
Interestingly, we depart from~\cite{Snoek2015} regarding the optimization of $\zb$. While their squared-loss formulation naturally lends itself to stochastic gradient descent (SGD), in a regime with moderate values of $T$ (typically several tens in our settings) the evidence~(\ref{eq:marginal_likelihood}) is better suited to batch optimization. In our experiments, L-BFGS~\cite{Byrd1995} worked well. Unlike~\cite{Snoek2015}, an important by-product of this choice is that we need not find hyperparameters for SGD that should work robustly across a broad set of BO problems.

\subsection{Random Fourier representation}
\label{sec:rf features}
An alternative approach is to use random kitchen sinks (RKS) for a random Fourier basis expansion~\cite{Rahimi2007}. Let $\Ub \in \Real^{D \times P}$ and $\bb \in \Real^{D}$ be such that $\Ub  \sim  \Ncal (\zerob,\Ib)$ and  $\{b_j \}_{j=1}^D \sim \Ucal([0,2 \pi ])$. For a vector $\xb$, RKS defines the mapping
$
\phib_\zb(\xb) = \sqrt{2/D} \cos ( \frac{1}{\sigma}  \Ub \xb   + \bb),
$
where $\sigma \in \Real^+$ is the bandwidth of the approximated RBF kernel. The parameter vector $\zb$ is a flattened version of $\{\Ub,\bb,\sigma\}$. Unlike the NN, the RKS representation contains only one hyperparameter to optimize ($\Ub$ and $\bb$ are randomly generated). This reduces the complexity of learning the map, but is less expressive as we show in the following section. To optimize $\sigma$, we proceed as for the weights of the NN (see Section~\ref{sec:nn features}).

\section{Results}
\label{sec:experiments}
The following subsections illustrate the benefits of multiple ABLR in a variety of settings. Sections \ref{sec:synthetic} and \ref{sec:openml} evaluate its ability to gather knowledge from multiple tasks, respectively on synthetic and \texttt{OpenML} data~\cite{Vanschoren2014}. Section \ref{sec:signals} shows how it can also be applied to exploit information from multiple heterogeneous signals. By doing so, we intend to learn more meaningful representations, which can be leveraged to accelerate the hyperparameter optimization. We could further generalize the model to handle multiple tasks and multiple signals at the same time, but leave this for future work.

We implemented multiple ABLR in \texttt{GPyOpt}~\cite{Gpyopt2016}, with a backend in \texttt{MxNet}~\cite{Chen2015}, fully benefiting from the symbolic computation to obtain the derivatives of the mappings  $\zb, \{\alpha_t, \beta_t \}_{t=1}^T \rightarrow \rho(\zb, \{\alpha_t, \beta_t \}_{t=1}^T) $, together with $\xb_t^* \rightarrow \mu_t(\xb_t^*; \Dcal_t, \alpha_t, \beta_t, \zb)$ and $\xb_t^* \rightarrow \sigma_t^2 (\xb_t^*; \Dcal_t, \alpha_t, \beta_t, \zb)$. In particular, we leverage the backward operator for the Cholesky decomposition~\cite{Seeger2017}. Interestingly, this allows us to jointly optimize all the model hyperparameters and perform exact BLR on top of an arbitrarily complex NN.

\subsection{Transfer learning across parametrized quadratic functions}
\label{sec:synthetic}
We first consider a set of $T$ tasks. A task takes the form of a parametrized 3-dimensional quadratic function
$
f_t(\xb) = \frac{1}{2}a_t \|\xb\|_2^2 + b_t \oneb^\top \xb + c_t ,
$
where $(a_t, b_t, c_t) \in [0.1,10]^3$. We call the triplet $(a_t, b_t, c_t)$ the context associated to each task $t$. In a real-world setting, the contextual information would correspond to meta-data, e.g., the data set size or its dimensionality, as we shall see in the next~section. We generated $T=30$ different tasks by drawing $(a_t, b_t, c_t)$ uniformly at random, and evaluated ABLR in a leave-one-task-out fashion. Specifically, we optimized each one of the 30 tasks after warm starting the optimization with 10 observations for the remaining 29 tasks. We compared single task ABLR-based and standard GP-based hyperparameter optimization (HPO), both denoted by \texttt{plain}, with their transfer learning counterparts, both denoted by \texttt{transfer}. We perform transfer learning with standard GPs by stacking all observations together and augmenting the input space with the corresponding contextual information~\cite{Krause2011}. For ABLR with transfer, we took our approach, i.e., one marginal likelihood per task, with and without the contextual information. 

\begin{figure*}[t]
\begin{subfigure}{.6\textwidth}
\centering
\includegraphics[width=0.95\textwidth]{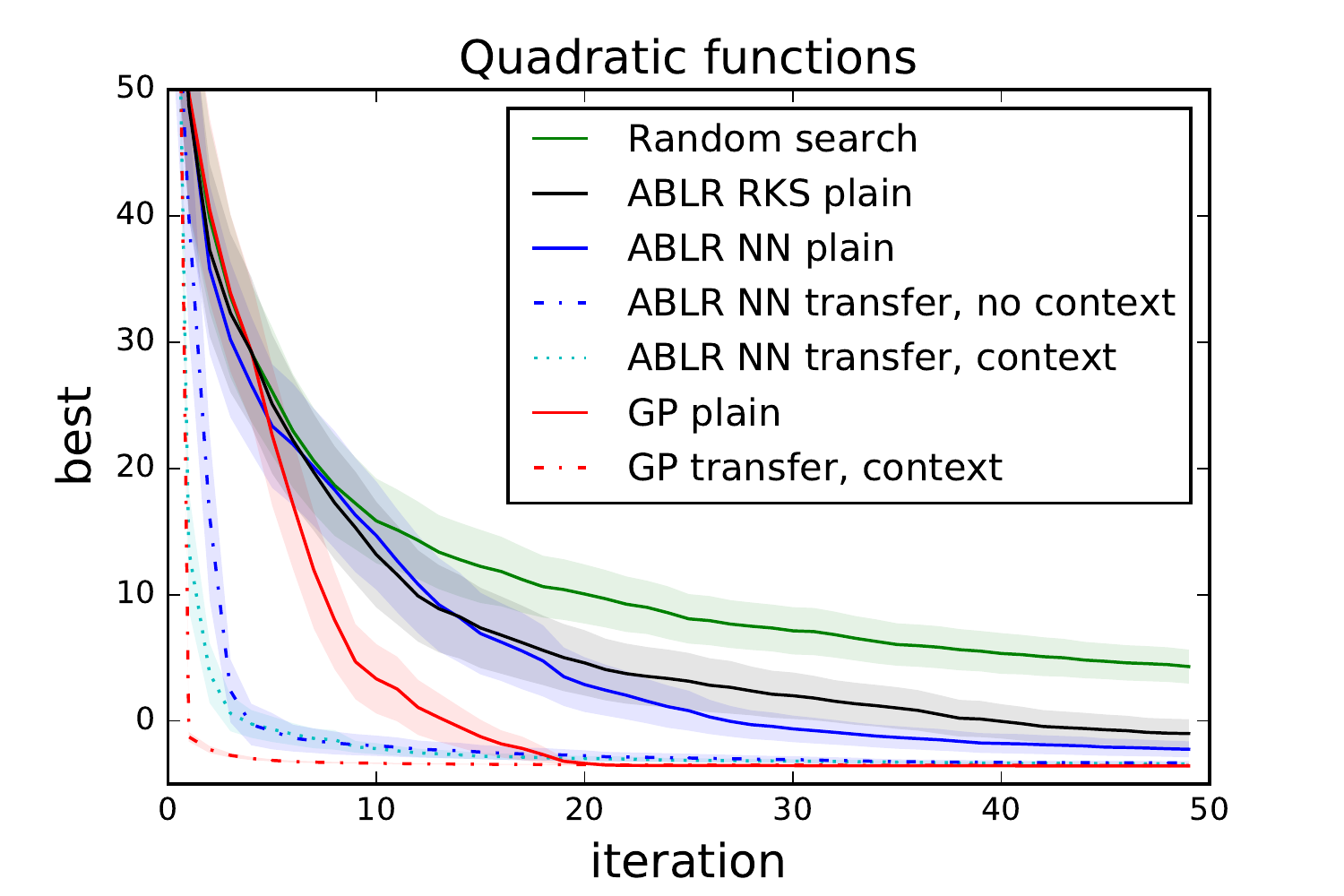} 
\end{subfigure}%
\begin{subfigure}{.4\textwidth}
\centering
\includegraphics[width=0.7\textwidth]{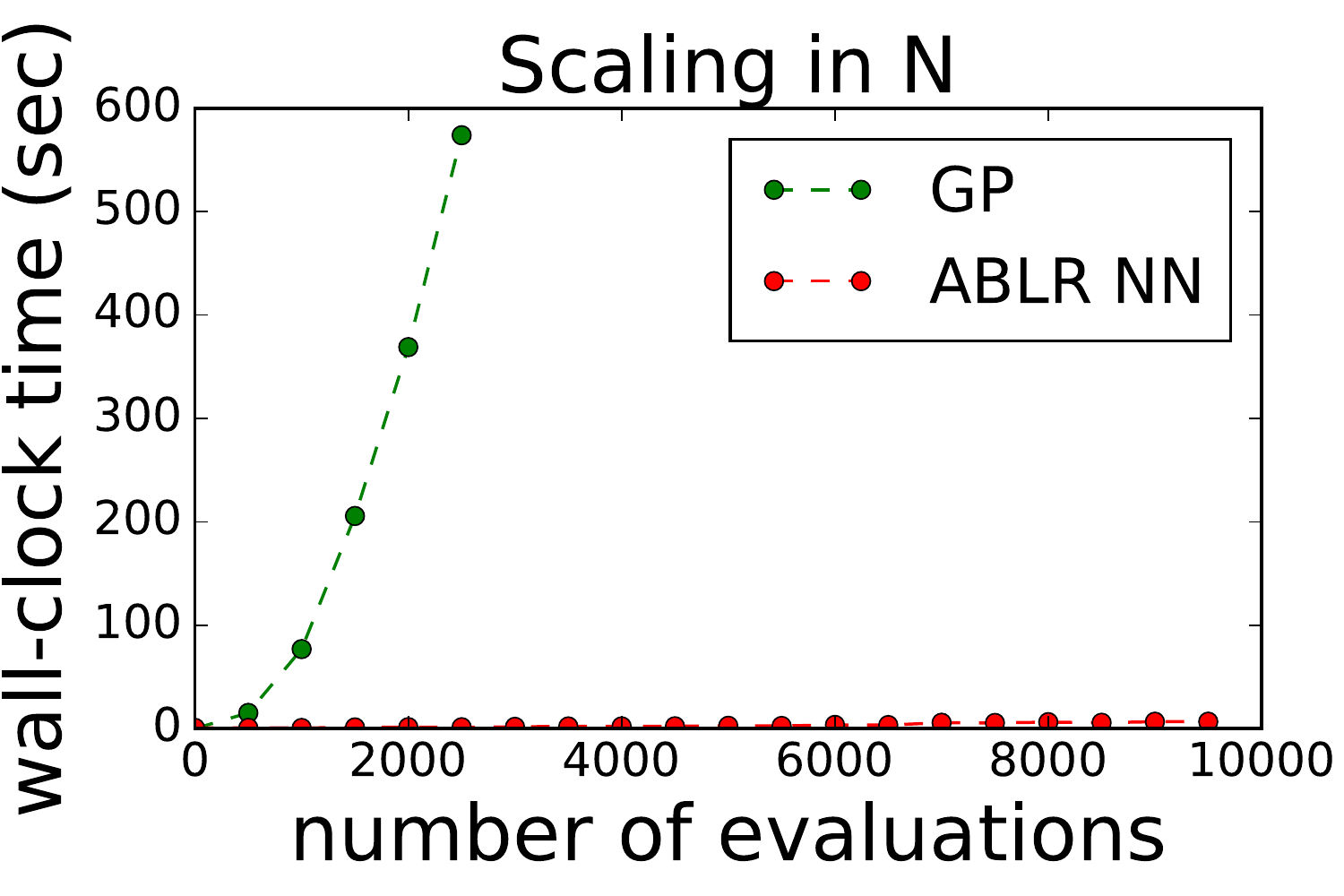}
\includegraphics[width=0.7\textwidth]{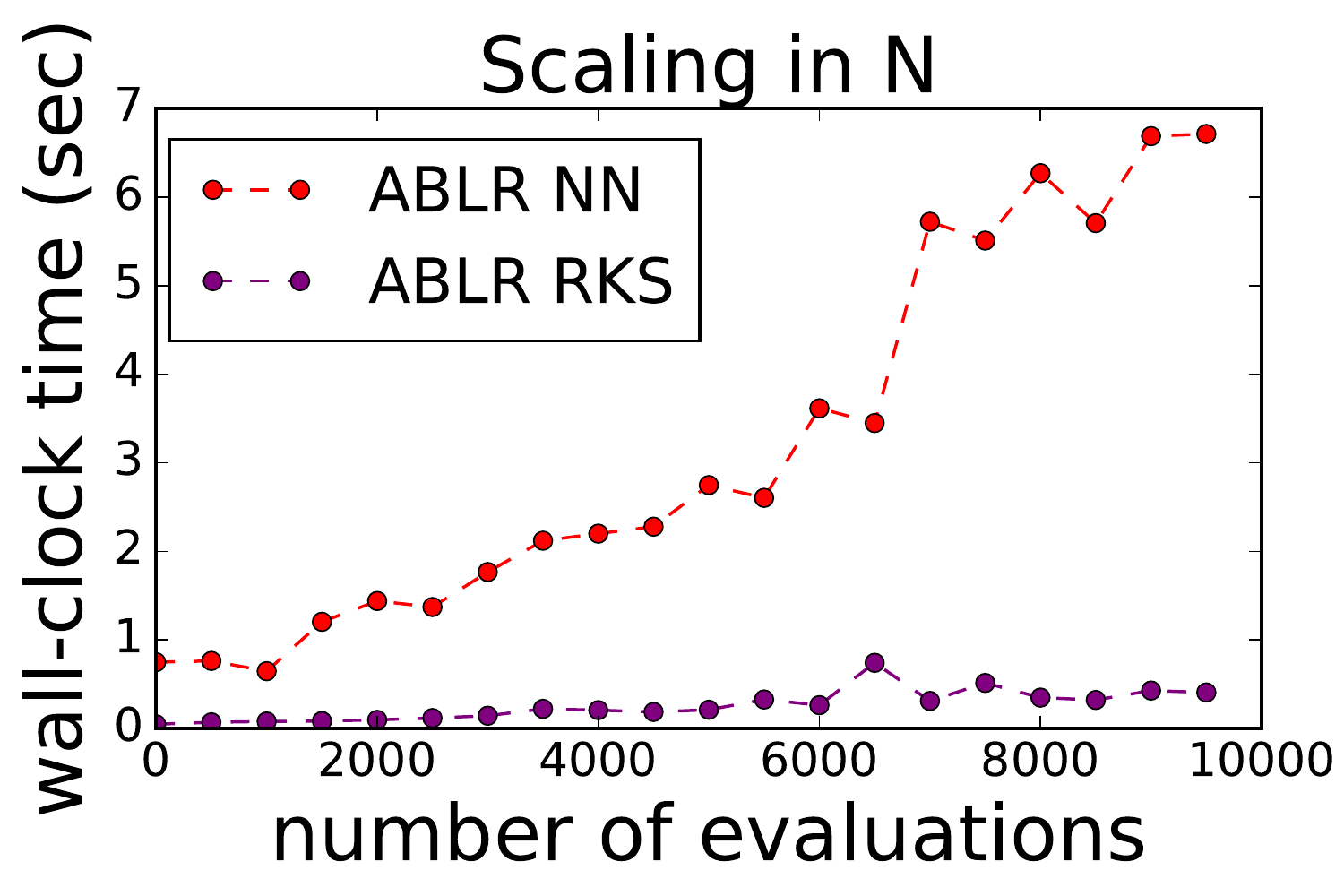} 
\end{subfigure}%
\caption { \textit{Left}: Transfer learning across parametrized quadratic functions. \textit{Right-top}: GP (cubical scaling) vs ABLR (linear scaling).   \textit{Right-bottom}: NN vs RKS basis expansion.}
\label{transfer_quadratic_GP_vs_ABLR}
\end{figure*}

Figure \ref{transfer_quadratic_GP_vs_ABLR}(left) shows the current best minimum at each of 50 iterations of HPO. The results are averaged over 10 random initializations and 30 leave-one-task-out runs.
HPO converges to the minimum much faster than plain ABLR or plain GP when we exploit the information from the related tasks. In addition, the RKS representation with $D=100$ performs slightly worse than the NN with 3 hidden layers of 50 units each per layer (as advocated in~\cite{Snoek2015}). Including the contextual information did not yield clear improvements, hence, for simplicity, we do not use it in the following experiments.
The GP-based HPO with transfer performs slightly better on this toy example, but is not applicable in large-scale settings, such as the one in the next section (with $\sum_t N_t \approx 7.5\times10^5$). Figure \ref{transfer_quadratic_GP_vs_ABLR}(right) compares the compute time of HPO with GP and NN-based ABLR, suggesting that the linear scaling with the number of evaluations of the latter allows us to apply ABLR in the large-scale setting. The RKS basis expansion further decreases the computational time (at the expense of performance).

\subsection{Transfer learning across OpenML black-box functions}
\label{sec:openml}
\begin{figure*}[h]
\begin{subfigure}{.5\textwidth}
\centering
\includegraphics[width=0.9\textwidth]{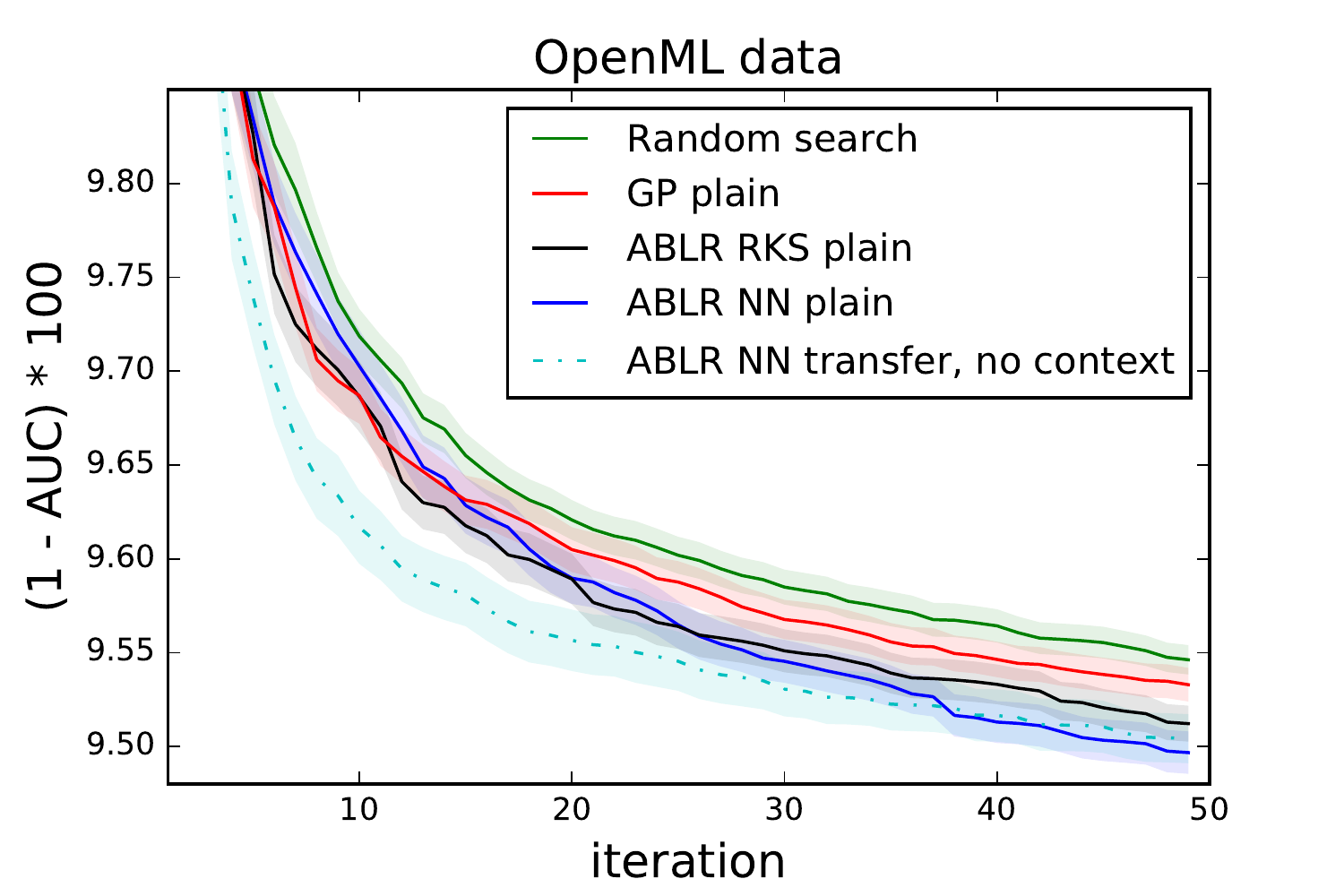}\vspace*{-0.2cm}
\end{subfigure}%
\begin{subfigure}{.5\textwidth}
\centering
\includegraphics[width=0.9\textwidth]{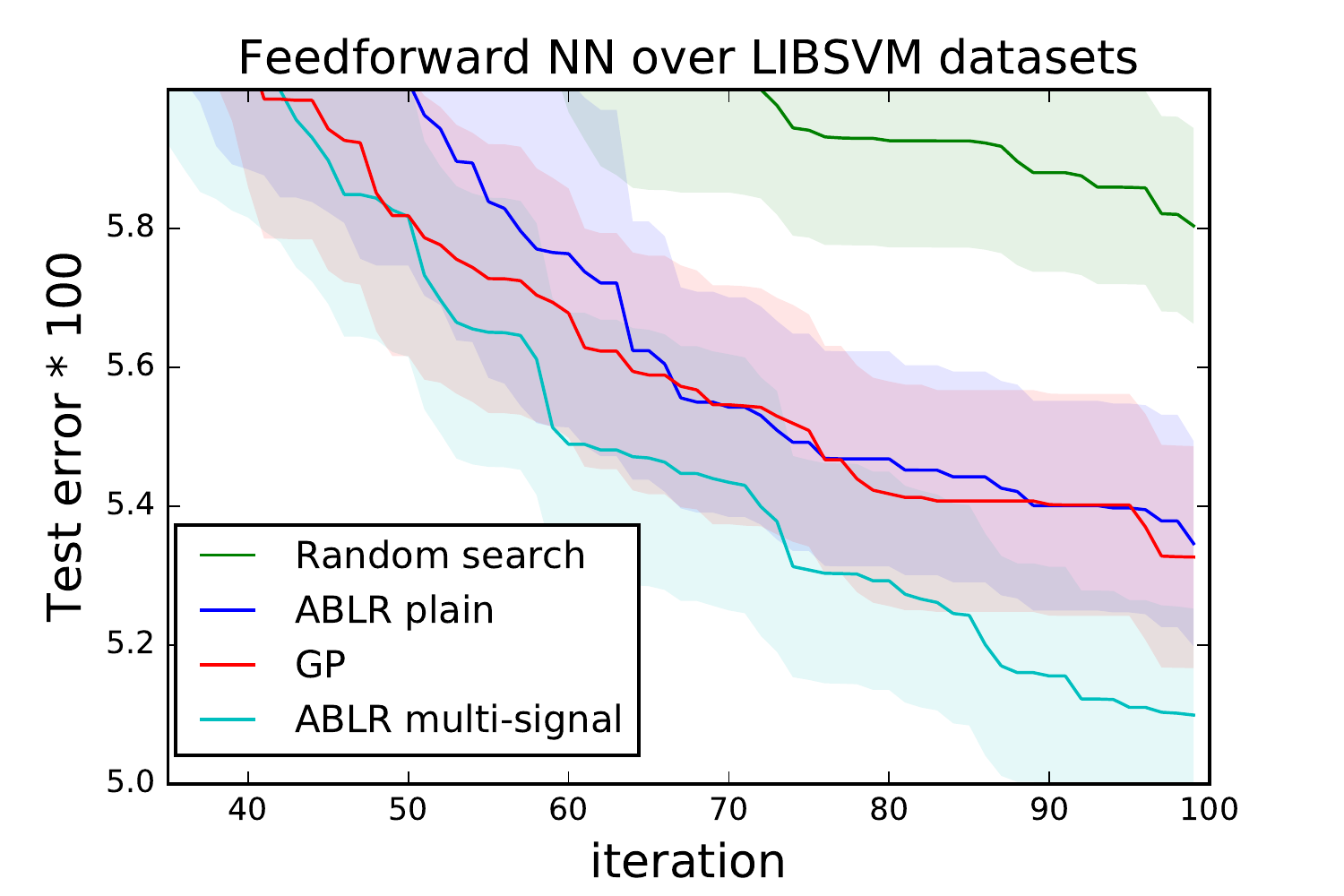}\vspace*{-0.2cm}
\end{subfigure}%
\caption {\textit{Left}: OpenML data, multiple tasks. \textit{Right}: LIBSVM data, multiple signals.}
\label{openml_libsvm_exp}
\end{figure*}

We consider the \texttt{OpenML} platform~\cite{Vanschoren2014}, which contains a large number of evaluations for a wide range of machine learning algorithms (referred to as flows in \texttt{OpenML}) over different datasets. In particular, we focus on a random forest model (\texttt{flow\_id} 6794) and apply ABLR to optimize its hyperparameters. We filtered the $T=30$ most evaluated datasets for this \texttt{flow\_id}, which amounts to $\sum_{t} N_t \approx 7.5\times10^5$ evaluations (with $N_t$ ranging from $9.940$ to $64.284$).
In this setting, the linear scaling of ABLR is particularly appealing. As previously, we apply a leave-one-task-out protocol, where each task stands for a dataset. For the left-out task being optimized, say $t_0$, we use the surrogate modeling approach from~\cite{Eggensperger2012}. We compare \texttt{GP plain} and \texttt{ABLR plain}, which use evaluations of task $t_0$ only, with \texttt{ABLR transfer}, which is warm-started with the evaluations of all the other tasks. The results are reported in Figure \ref{openml_libsvm_exp}(left), showing that ABLR is able to gather knowledge from different datasets to speed up the convergence. 

\subsection{Tuning of feedforward neural networks from heterogeneous signals}
\label{sec:signals}
Finally, we consider the tuning of feedforward NNs for binary classification. We show that our formulation can be seamlessly applied to the orthogonal problem of modeling $S$ multiple output signals, possibly of heterogeneous nature, \textit{at once}. Here, we optimize for the validation accuracy, using the training accuracy and CPU time as side information. Such side signals ``come for free'' while training machine learning algorithms, but are in general not exploited for efficient HPO. In comparison to multi-output GPs that scale as $\Ocal(N^3+S^3)$, ABLR scales as $\Ocal(S(D^2N + D^3))$.
The NN hyperparameters to tune are the number of hidden layers in $\{1,\dots,4\}$, the number of units in $\{1,\dots,50\}$, the amount of $\ell_2$ regularization in $\{2^{-6},2^{-5},\dots,2^3\}$, the learning rate of Adam~\cite{Kingma2014} in $\{2^{-6},2^{-5},\dots,2^{-1}\}$, and the number of epochs in $\{3,\dots,10\}$. Figure~\ref{openml_libsvm_exp}(right) shows the results, which are averaged over 10 random initializations and 5 datasets (\texttt{w8a, sonar, w1a, phishing, australian}) from LIBSVM~\cite{Chang2011}.
It can be observed that incorporating side signals in addition to the target signal, namely the validation accuracy of the NN classifier, speeds up the ABLR-based HPO.

\bibliographystyle{abbrv}

\begin{thebibliography}{10}

\bibitem{Gpyopt2016}
\texttt{GPyOpt}: A {B}ayesian optimization framework in python.
\newblock \texttt{http://github.com/SheffieldML/GPyOpt}, 2016.

\bibitem{Bardenet2013}
R.~Bardenet, M.~Brendel, B.~K{\'e}gl, and M.~Sebag.
\newblock Collaborative hyperparameter tuning.
\newblock In {\em Proceedings of the International Conference on Machine
  Learning (ICML)}, pages 199--207, 2013.

\bibitem{Bishop2006}
C.~M. Bishop.
\newblock {\em {Pattern Recognition and Machine Learning}}.
\newblock Springer New York, 2006.

\bibitem{Byrd1995}
R.~H. Byrd, P.~Lu, J.~Nocedal, and C.~Zhu.
\newblock A limited memory algorithm for bound constrained optimization.
\newblock {\em SIAM Journal on Scientific Computing}, 16(5):1190--1208, 1995.

\bibitem{Chang2011}
C.-C. Chang and C.-J. Lin.
\newblock {LIBSVM}: A library for support vector machines.
\newblock {\em ACM Transactions on Intelligent Systems and Technology},
  2:27:1--27:27, 2011.

\bibitem{Chen2015}
T.~Chen, M.~Li, Y.~Li, M.~Lin, N.~Wang, M.~Wang, T.~Xiao, B.~Xu, C.~Zhang, and
  Z.~Zhang.
\newblock Mxnet: A flexible and efficient machine learning library for
  heterogeneous distributed systems.
\newblock In {\em Neural Information Processing Systems, Workshop on Machine
  Learning Systems}, 2015.

\bibitem{Eggensperger2012}
K.~Eggensperger, F.~Hutter, H.~Hoos, and K.~Leyton-brown.
\newblock Efficient benchmarking of hyperparameter optimizers via surrogates
  background: hyperparameter optimization.
\newblock In {\em Proceedings of the 29th AAAI Conference on Artificial
  Intelligence}, pages 1114--1120, 2012.

\bibitem{Feurer2015}
M.~Feurer, T.~Springenberg, and F.~Hutter.
\newblock Initializing {B}ayesian hyperparameter optimization via
  meta-learning.
\newblock In {\em Proceedings of the Twenty-Ninth AAAI Conference on Artificial
  Intelligence}, 2015.

\bibitem{Fusi2017}
N.~Fusi and H.~M. Elibol.
\newblock Probabilistic matrix factorization for automated machine learning.
\newblock Technical report, preprint arXiv:1705.05355, 2017.

\bibitem{Golovin2017}
D.~Golovin, B.~Solnik, S.~Moitra, G.~Kochanski, J.~Karro, and D.~Sculley.
\newblock Google vizier: A service for black-box optimization.
\newblock In {\em Proceedings of the 23rd ACM SIGKDD International Conference
  on Knowledge Discovery and Data Mining}, pages 1487--1495, 2017.

\bibitem{Hernandez-Lobato2017}
J.~M. Hern{\'a}ndez-Lobato, J.~Requeima, E.~O. Pyzer-Knapp, and
  A.~Aspuru-Guzik.
\newblock Parallel and distributed {T}hompson sampling for large-scale
  accelerated exploration of chemical space.
\newblock In {\em Proceedings of the International Conference on Machine
  Learning (ICML)}, 2017.

\bibitem{Jenatton2017b}
R.~Jenatton, C.~Archambeau, J.~Gonzales, and M.~Seeger.
\newblock Bayesian optimization with tree-structured dependencies.
\newblock In {\em Proceedings of the International Conference on Machine
  Learning (ICML)}, 2017.

\bibitem{Kingma2014}
D.~Kingma and J.~Ba.
\newblock Adam: A method for stochastic optimization.
\newblock Technical report, preprint arXiv:1412.6980, 2014.

\bibitem{Krause2011}
A.~Krause and C.~S. Ong.
\newblock Contextual gaussian process bandit optimization.
\newblock In {\em Advances in Neural Information Processing Systems (NIPS)},
  pages 2447--2455, 2011.

\bibitem{MacKay2003}
D.~J.~C. Mackay.
\newblock {\em Information Theory, Inference and Learning Algorithms}.
\newblock Cambridge University Press, 2003.

\bibitem{McIntire2016}
M.~McIntire, D.~Ratner, and S.~Ermon.
\newblock Sparse gaussian processes for {B}ayesian optimization.
\newblock In {\em Proceedings of the Conference on Uncertainty in Artificial
  Intelligence (UAI)}, 2016.

\bibitem{Rahimi2007}
A.~Rahimi, B.~Recht, et~al.
\newblock Random features for large-scale kernel machines.
\newblock In {\em Advances in Neural Information Processing Systems (NIPS)},
  volume~3, page~5, 2007.

\bibitem{Rasmussen2006}
C.~Rasmussen and C.~Williams.
\newblock {\em Gaussian Processes for Machine Learning}.
\newblock MIT Press, 2006.

\bibitem{Seeger2017}
M.~Seeger, A.~Hetzel, Z.~Dai, and N.~D. Lawrence.
\newblock Auto-differentiating linear algebra.
\newblock Technical report, preprint arXiv:1710.08717, 2017.

\bibitem{Shahriari2016}
B.~Shahriari, K.~Swersky, Z.~Wang, R.~P. Adams, and N.~de~Freitas.
\newblock Taking the human out of the loop: A review of {B}ayesian
  optimization.
\newblock {\em Proceedings of the IEEE}, 104(1):148--175, 2016.

\bibitem{Snoek2015}
J.~Snoek, O.~Rippel, K.~Swersky, R.~Kiros, N.~Satish, N.~Sundaram, M.~Patwary,
  M.~Prabhat, and R.~Adams.
\newblock Scalable {B}ayesian optimization using deep neural networks.
\newblock In {\em Proceedings of the International Conference on Machine
  Learning (ICML)}, pages 2171--2180, 2015.

\bibitem{Springenberg2016}
J.~T. Springenberg, A.~Klein, S.~Falkner, and F.~Hutter.
\newblock Bayesian optimization with robust {B}ayesian neural networks.
\newblock In {\em Advances in Neural Information Processing Systems (NIPS)},
  pages 4134--4142, 2016.

\bibitem{Swersky2013}
K.~Swersky, J.~Snoek, and R.~P. Adams.
\newblock Multi-task {B}ayesian optimization.
\newblock In {\em Advances in Neural Information Processing Systems (NIPS)},
  pages 2004--2012, 2013.

\bibitem{Vanschoren2014}
J.~Vanschoren, J.~N. Van~Rijn, B.~Bischl, and L.~Torgo.
\newblock {OpenML}: networked science in machine learning.
\newblock {\em ACM SIGKDD Explorations Newsletter}, 15(2):49--60, 2014.

\bibitem{Yogatama2014}
D.~Yogatama and G.~Mann.
\newblock Efficient transfer learning method for automatic hyperparameter
  tuning.
\newblock In {\em Proceedings of the International Conference on Artificial
  Intelligence and Statistics (AISTATS)}, pages 1077--1085, 2014.

\end{thebibliography}

\end{document}